\documentclass[letterpaper, 10 pt, conference]{ieeeconf}  

\usepackage{amsmath,graphicx}
\usepackage{multirow}
\usepackage{amsmath}
\usepackage{amssymb}
\usepackage{cite}
\usepackage{colortbl}
\usepackage{xcolor}
\usepackage[pagebackref=true,breaklinks=true,colorlinks,bookmarks=false]{hyperref}

\usepackage{color}

\IEEEoverridecommandlockouts                              

\title{\LARGE \bf QUEST: Query Stream for Practical Cooperative Perception}

\author{Siqi Fan\textsuperscript{1}, Haibao Yu\textsuperscript{2,1}, Wenxian Yang\textsuperscript{1}, Jirui Yuan\textsuperscript{1}, Zaiqing Nie\textsuperscript{1*}\\
\textsuperscript{1}Institute for AI Industry Research (AIR), Tsinghua University  \textsuperscript{2} The University of Hong Kong\
}

\begin{document}

\maketitle

\renewcommand{\thefootnote}{\fnsymbol{footnote}}
\footnotetext[1]{To whom the correspondence should be addressed. For any questions or discussions, please email \{fansiqi, dair\}@air.tsinghua.edu.cn.}

\thispagestyle{empty}
\pagestyle{empty}

\begin{abstract}
  Cooperative perception can effectively enhance individual perception performance by providing additional viewpoint and expanding the sensing field. Existing cooperation paradigms are either interpretable (result cooperation) or flexible (feature cooperation). In this paper, we propose the concept of query cooperation to enable interpretable instance-level flexible feature interaction. To specifically explain the concept, we propose a cooperative perception framework, termed QUEST, which let query stream flow among agents. The cross-agent queries are interacted via fusion for co-aware instances and complementation for individual unaware instances. Taking camera-based vehicle-infrastructure perception as a typical practical application scene, the experimental results on the real-world dataset, DAIR-V2X-Seq, demonstrate the effectiveness of QUEST and further reveal the advantage of the query cooperation paradigm on transmission flexibility and robustness to packet dropout. We hope our work can further facilitate the cross-agent representation interaction for better cooperative perception in practice.
\end{abstract}


\section{Introduction}

Despite the significant progress have been made in individual perception, intelligent vehicles still have to face challenges of unobservable dangers caused by occlusion and limited perception range. Different from the individual perception which senses the surrounding with its own onboard sensor system, cooperative perception, especially vehicle-infrastructure cooperative perception (VICP), shed light on reliable autonomous driving in a complex traffic environment and have achieved increasing attention recently \cite{9732063, han2023collaborative}. Leveraging the roadside sensor system with more flexible mounting height and posture, the cooperative perception field is effectively extended, and some challenging individual perception cases (e.g., long-range small object detection) can be readily tackled in VICP setting \cite{yu2022dair, cbr}.

\begin{figure}[t]
  \centering
  \includegraphics[scale=0.45]{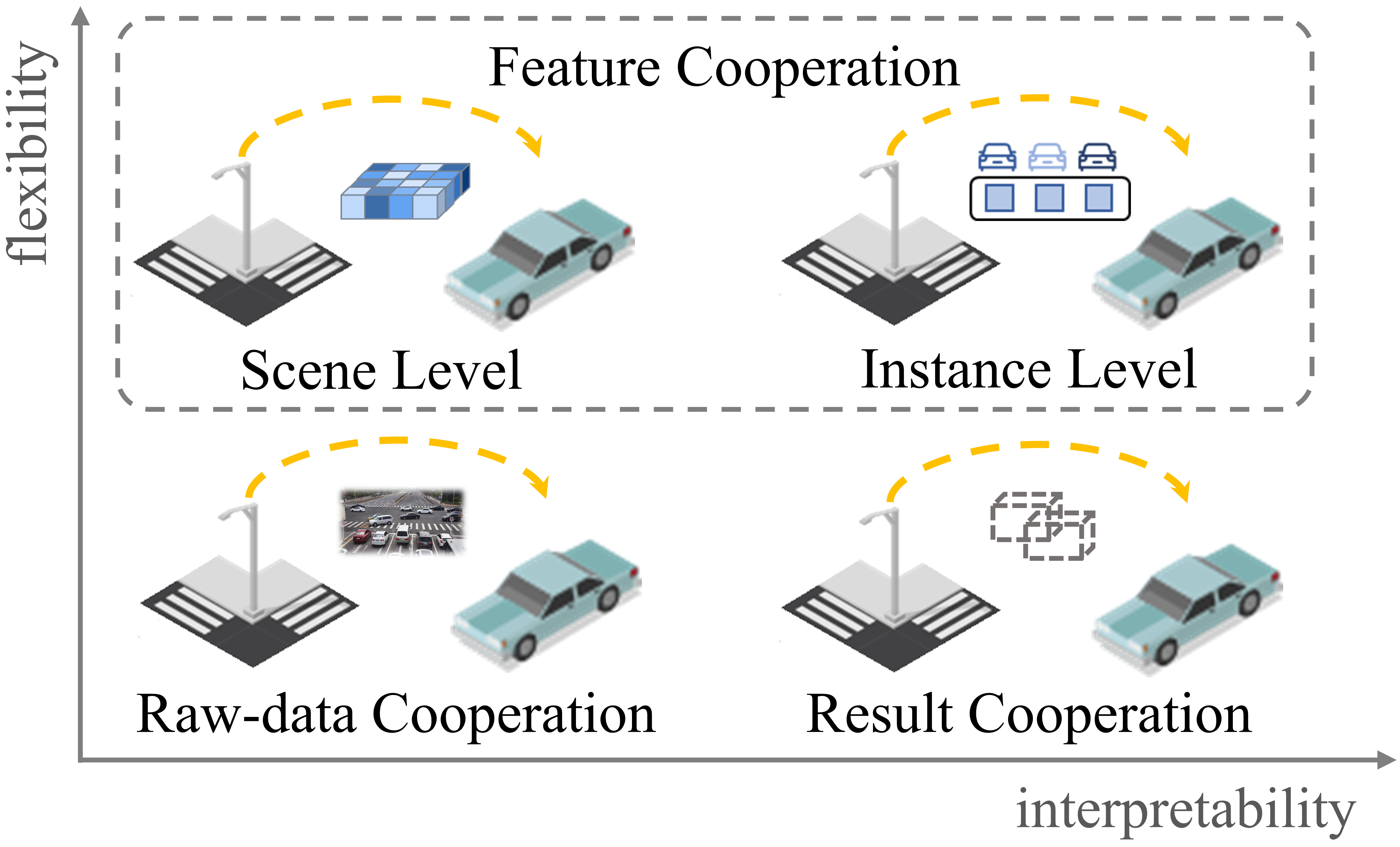}
  \caption{Query cooperation enables instance-level feature cooperation, which is more interpretable than scene-level feature cooperation and more flexible than instance-level result cooperation.}
  \label{fig:query-coop}
\end{figure}

Advantages are usually followed by new challenges. Naturally, the first and foremost question is how to cooperate between multiple agents. According to what is shared among agents, there are three typical cooperation paradigms \cite{9732063, bai2022survey, han2023collaborative}, including data cooperation (early fusion), feature cooperation (intermediate fusion), and result cooperation (late fusion). Data cooperation \cite{9228884, cooper} is regarded as the upper bound of performance since the comprehensive information is interchanged along with raw data across agents. However, the high transmission cost of massive data is unbearable in practical applications. Result cooperation is widely deployed in practice due to the advantages of bandwidth-economic, where agents only share predictions \cite{9228884, yu2022dair}. Nevertheless, the significant information loss in result cooperation makes it highly reliant on accurate individual predictions. Compared with those two paradigms, feature cooperation \cite{fcooper, cui2022coopernaut, when2com, wang2020v2vnet, xu2022v2x, wang2023vimi, coca3d, ffnet} is more flexible and performance-bandwidth balanced, as the information loss is controllable via feature selection and compression. Even though some of them have achieved region-level feature selection \cite{where2comm}, the interpretability of feature selection and fusion are still limited, since the scene-level features abstractly represent the whole observable region. It is worth noting that the interaction between predictions in result cooperation is instance-level, resulting in physically interpretable cooperation targets.


Addressing that, we naturally come up with a question: \textit{is there an eclectic approach for cooperative perception, which is both interpretable and flexible?}

Inspired by the success of transformer-based methods in individual perception tasks\cite{detr, detr3d, motr}, we propose the concept of query cooperation, which is an instance-level feature interaction paradigm based on the query stream across agents, standing on the midpoint between scene-level feature cooperation and instance-level result cooperation (Figure~\ref{fig:query-coop}). The instance-level cooperation makes it more physically interpretable, and feature interaction introduces more information elasticity. Specifically, we propose a framework, named QUEST, as a representative approach to describe the concept, where queries flow in the stream among agents. Firstly, each agent performs individual transformer-based perception. Every query output from the decoder corresponds to a possible detected object, and the query will be shared if its confidence score meets the requirement of the request agent. As the cross-agent queries arrive, they are utilized for both query fusion and complementation. Theoretically, query fusion can enhance the feature of the sensed instance with the feature from other viewpoints, while query complementation can directly complement the unaware instance of the local perception system. Then, the queries are used for cooperative perception, resulting in the final perception results. To evaluate the performance of QUEST, we generate the camera-centric cooperation labels on DAIR-V2X-Seq based on the single-side groundtruth labeled at the image-captured timestamps \textsuperscript{*}.
\footnotetext[1]{The original cooperation groundtruth is labeled at LiDAR's timestamp \cite{yu2022dair}, which is not suitable for camera-based researches.}

Our contributions are summarized as follows:
\begin{itemize}
  \item We propose the concept of query cooperation paradigm for cooperative perception task, which is more interpretable than scene-level feature cooperation and more flexible than result cooperation.
  \item A query cooperation framework, termed QUEST, is proposed as a representative approach. The cross-agent queries interact at the instance level via fusion and complementation.
  \item We take the camera-based vehicle-infrastructure cooperative object detection as a typical application scene. The experimental results on the real-world dataset, DAIR-V2X-Seq, demonstrate the effectiveness of QUEST and further show the advantage of the query cooperation paradigm on flexibility and robustness. Besides, the camera-centric cooperation labels are generated to facilitate the further development of the related researches.
\end{itemize}

\begin{figure*}
  \centering
  \includegraphics[scale=0.44]{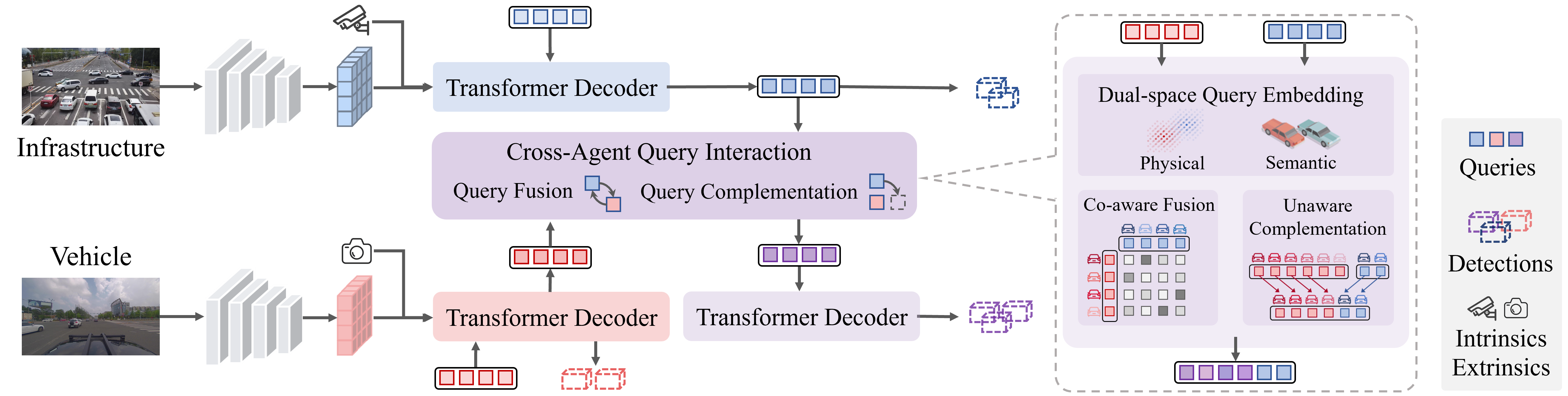}
  \caption{Architecture of QUEST framework.}
  \label{fig:quest}
\end{figure*}

\section{Related Works}

In this section, we briefly review two related topics, cooperative perception and query-based perception.

\subsection{Cooperative perception}

To break the sensing range limitation of onboard sensor systems and eliminate the influences of unobservable dangers, cooperative perception 
has attracted increasing attention in recent years. The most intuitive approach is data cooperation, which transmits raw sensor data and fundamentally overcomes the occlusion and long-range perception problem. Since 3D data can be directly aggregated, most data cooperation approaches are LiDAR-based \cite{9228884, cooper}. Although raw data reserves comprehensive information, the high transmission cost makes it challenging to deploy in practice. For the convenience of communication, result cooperation only transmits perception predictions, which is the most bandwidth-economic \cite{9228884, yu2022dair}. In addition, the instance-level bounding box aggregation makes the cooperation more physically interpretable. However, the performance of result cooperation highly relies on the accurate individual perception and precise parameters for coordinate system transformation. Therefore, recent methods pay more attention to feature cooperation, which can achieve better performance-bandwidth balance \cite{fcooper, cui2022coopernaut, when2com, wang2020v2vnet, xu2022v2x, where2comm, wang2023vimi, coca3d, ffnet}. Compared with the simple bounding box, the feature map is more flexible for both fusion and compression, but the scene-level feature cooperation is redundant for object perception and less explainable. Aiming on interpretable flexible cooperation, we propose the concept of query cooperation, which transmits instance-level features across agents. 

\subsection{Query-based perception}

Since the pioneering work DETR \cite{detr} is proposed for 2D object detection, the object query has been adopted for more and more perception tasks, including 3D detection and tracking. Query-based methods typically utilize sparse learnable queries for attentive feature aggregation. DETR3D \cite{detr3d} predicts 3D locations of queries and obtains the corresponding image features via projection. PETR \cite{petr} turns to embed image features with 3D position and directly learns the mapping relations using the attention mechanism. BEVFormer \cite{bevformer, bevformerv2} tackles the perception from a bird-eye view with grid-shaped queries and manages to realize spatial-temporal feature interaction through the deformable transformer. Leveraging temporal information, query-based methods are also beneficial to object tracking. To model cross-frame object association, MOTR \cite{motr} and TrackFormer \cite{trackformer} propose track query based on single frame object query. MUTR \cite{mutr} and PF-Track \cite{pftr} utilizes track query and achieve promising tracking performance for multi-view tasks. All of the existing query-based methods are developed for individual perception, we further extend it to cooperative perception in this paper. Specifically, we propose the QUEST framework to achieve a query stream across agents and design the cross-agent query interaction module for query fusion and complementation.

\section{Query Cooperation Paradigm}

What to share and how to cooperate are the two main concerns for practical cooperative perception, especially considering the limited bandwidth of the wireless communication. To design a better cooperation strategy, it is expected to be both interpretable and flexible, since interpretability leads to controllable cooperation and flexibility provides more operation space and possibilities. Considering that, we propose the query cooperation paradigm, which shares features across agents and performs cooperation via instance-level feature interaction. 

For clarity, we take vehicle-infrastructure cooperative perception as an example.

\textbf{Query Generation.} Both vehicle and infrastructure perform individual perception all the time, and each perception prediction $\mathcal{P}$ is corresponded to an object query $\mathcal{Q}$, according to the theory of transformer-based perception,
\begin{equation}
  \mathcal{P} = g(\mathcal{Q}) = g(f(\mathcal{D}))
\end{equation},
where $f(\cdot)$ is the feature extraction function for queries, $g(\cdot)$ is the query-based prediction function, and $\mathcal{D}$ denotes the input sensor data. 

\textbf{Query Transmission.} The query cooperation is triggered when the vehicle requests additional information from infrastructure side. Noting that the query request can be along with a specific instance-level requirement, like confidence threshold and region mask. Then, the queries met the requirement $\mathcal{Q}_{inf}$ are posted to the vehicle side. 

\textbf{Query Interaction.} Both the received queries $\mathcal{Q}_{inf}$ and local queries $\mathcal{Q}_{veh}$ are leveraged for further cooperative perception, and the query interaction strategy is to determine how to enhance and complement the $\mathcal{Q}_{veh}$ with $\mathcal{Q}_{inf}$.
\begin{equation}
  \mathcal{Q}_{coop} = h(\mathcal{Q}_{veh}, \mathcal{Q}_{inf})
\end{equation},
where $h(\cdot)$ denotes the query interaction function and $\mathcal{Q}_{coop}$ is the generated cooperative query set. 

\textbf{Query-based Prediction}. $\mathcal{Q}_{coop}$ is further fed into query-based prediction heads for perception tasks, resulting in the final cooperative perception predictions $\mathcal{P}_{coop}$.
\begin{equation}
  \mathcal{P}_{coop} = g(\mathcal{Q}_{coop}).
\end{equation}

\section{QUEST Framework}

To elaborate on the concept of query cooperation, we describe the proposed representative framework in this section. Benefiting from the deployment convenience, camera-based sensor systems are widely adopted in practical applications. Thus, we take the camera-based vehicle-infrastructure cooperative perception as a typical scenario to describe the framework.

\subsection{Overall Architecture}

As illustrated in Figure~\ref{fig:quest}, QUEST achieves cooperative perception via a cross-agent QUEry STream. The object queries flow from the infrastructure side to the vehicle side when query cooperation is triggered by the vehicle. The framework mainly consists of two functional modules, including single-agent query-based perception modules and a cross-agent query interaction module.

For every single agent, like the vehicle, the query-based perception module is continuously running to ensure the basic individual perception capability, leveraging its own sensor data obtained from the onboard system. It will always output perception predictions whether the query cooperation is triggered or not. Theoretically, every query-based perception method can be directly plugged in, and we adopt PETR \cite{petr} as an example in this paper. The captured image is fed into the backbone for feature extraction, and both the feature and calibration parameters are input to a transformer-based decoder to perform object detection. Each prediction is matched with a corresponding object query, and it is the source of the query stream. Considering the limited bandwidth of wireless communication, the infrastructure-side query stream is shunted according to a confidence score threshold required by the vehicle side, resulting in a high-quality sparse feature transmission.

When the infrastructure-side query stream flows to the vehicle side, it joins the local query stream to form a cooperative query stream. The cross-agent query interaction module is designed to integrate the object queries from different sources, which is elaborated in the following subsection. The joint query stream finally flocks to the transformer-based decoder, and the cooperative predictions are output.

\subsection{Cross-agent Query Interaction}

Similar to all the other cooperation paradigms, how to aggregate the cross-agent information is always the most important part of the framework. Benefiting from the interpretable instance-level cooperation, the query interaction mechanism is natural, including query fusion for co-aware objects and query complementation for unaware objects. 

In the first place, the corresponding location of the cross-agent queries should be transformed into a unified coordinate system, which is generally the vehicle-side LiDAR coordinate system. Since each query is along with a 3D reference point, the transformation is readily performed using the calibration parameters (rotation and translation matrix).

The instance-level predictions are matched according to their locations in result cooperation. Although the strategy can be directly adopted in QUEST, it relies on both the accurate location prediction and precise coordinate transformation. To realize more robust query matching, we propose the dual-space query embedding.

\begin{figure}[h]
  \centering
  \includegraphics[scale=0.65]{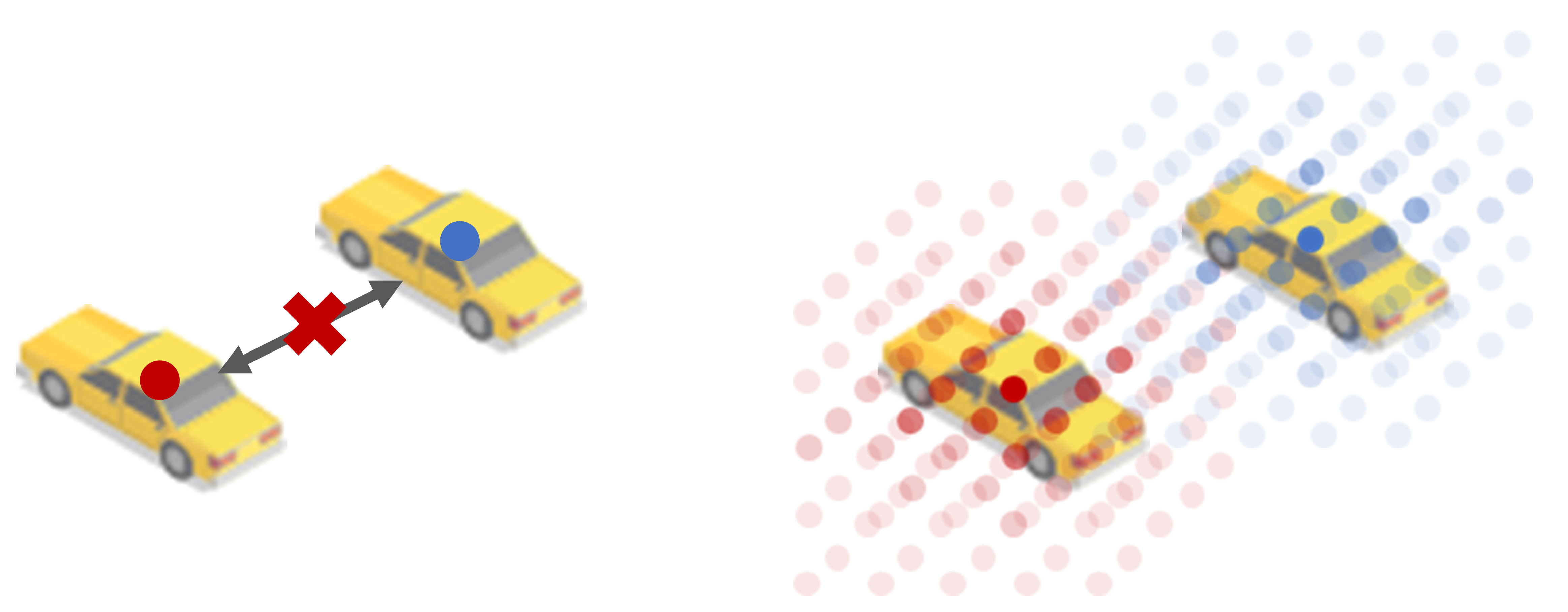}
  \caption{Illustration of the location grid for dual-space query embedding. Compared with the exact center-based matching, grid-based matching is more robust with location noise.}
  \label{fig:dse}
\end{figure}

\textbf{Dual-space Query Embedding} takes both location information and semantic information into consideration, which is embedded in physical and feature space. For location embedding, we expand the exact center to a grid to give a high tolerance of location noise, as shown in Figure~\ref{fig:dse}. The 3D coordinates in the grid are concatenated to form grid embedding after normalization. However, the loose constraint of location will inevitably introduce false-matched pairs. We further take semantic information into account to pay additional attention to appearance. Technically, the query’s feature is concatenated with the grid embedding $\mathcal{E}_g$, and the dual-space query embedding $\mathcal{\hat{E}}$ is generated using a multi-layer perceptron (MLP) encoder.
\begin{equation}
    \mathcal{\hat{E}} = \mathbf{MLP}(\mathcal{E}_g \oplus \mathcal{E}_f)
\end{equation}
, where $\oplus$ is the concatenation operation, $\mathbf{MLP}$ denotes the multi-layer perceptron encoder, and $\mathcal{E}_f$ is the semantic embedding. We directly regard the query's feature as semantic embedding in this work.

\textbf{Cross-agent Query Alignment} is a specific and necessary operation for query cooperation, which is mainly due to the implicit encoding of the instance-level orientation. The prediction's orientation is explicitly represented in result cooperation, and the orientation of the dense feature map is directly related to the corresponding coordinate system. Therefore, both of them can achieve orientation transformation via explicit coordinate system transformation. However, the implicit encoded feature in instance-level query can not be manually operated, even if the orientation-related feature is decoupled from others. We adopt MLP for feature space alignment, which enables implicit orientation transformation and cross-agent feature alignment.
\begin{equation}
    \mathcal{\hat{Q}}_{inf} = \mathbf{MLP}(\mathcal{Q}_{inf} \oplus \mathcal{R}_{I2V})
\end{equation}
, where $\mathcal{Q}_{inf}$ is the infrastructure-side query, and $\mathcal{R}_{I2V}$ is the rotation matrix from infrastructure side to vehicle side.

\textbf{Attentive Query Fusion} is to enhance the vehicle-side aware queries with the queries from the infrastructure-side view. The fusion is attentively guided by the dual-space query embedding. Specifically, we calculate the embedding distance between each two query pairs and generate the attentive fusion weights on the basis of that via MLP. Take the $i-th$ vehicle-side query $\mathcal{Q}_{veh}^{i}$ and the $j-th$ infrastructure-side query $\mathcal{\hat{Q}}_{inf}^{j}$ as an example,
\begin{equation}
    \mathcal{W}_{i,j} = \mathbf{MLP}(||\mathcal{\hat{E}}_{veh}^{i} - \mathcal{\hat{E}}_{inf}^{j}||_{2}) 
\end{equation}
, where $\mathcal{\hat{E}}_{veh}^{i}$ and $\mathcal{\hat{E}}_{inf}^{j}$ denote the generated dual-space query embedding, and $||\cdot||_{2}$ is the $L_{2}$ distance function. Then, the vehicle-side query stream is updated and formed to the cooperative query stream $Q_{coop}$ via weighted summation.
\begin{equation}
    \mathcal{Q}_{coop}^{i} = \mathcal{Q}_{veh}^{i} + \mathcal{W}_{i,j} * \mathcal{\hat{Q}}_{inf}^{j}
\end{equation}

\begin{figure}[t]
  \centering
  \includegraphics[scale=0.95]{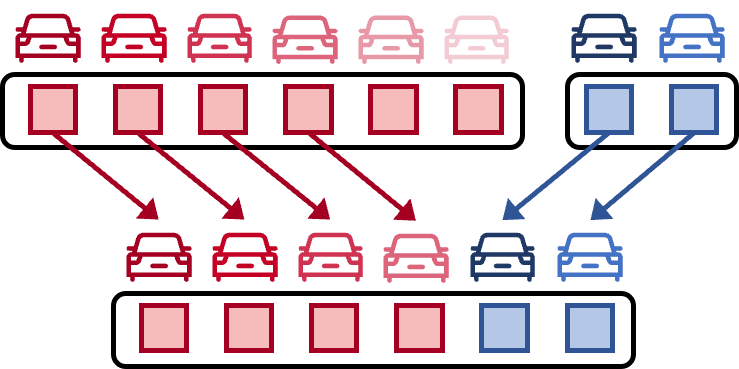}
  \caption{Illustration of the cross-agent query complementation. The local queries with low confidence scores are replaced with the received queries to reduce additional computational costs.}
  \label{fig:query-compensation}
\end{figure}

\textbf{Query Complementation} is to complement the vehicle-side unaware object queries with the received infrastructure-side queries. Instead of simply inserting the cross-agent queries into the local query stream, we turn to a replacement strategy to reduce the extra computational cost. Firstly, the vehicle-side query is sorted according to the confidence score. The received queries are then used to replace the queries with low confidence scores, as shown in Figure~\ref{fig:query-compensation}.

\section{Experiments}

\begin{figure*}
  \centering
  \includegraphics[scale=0.52]{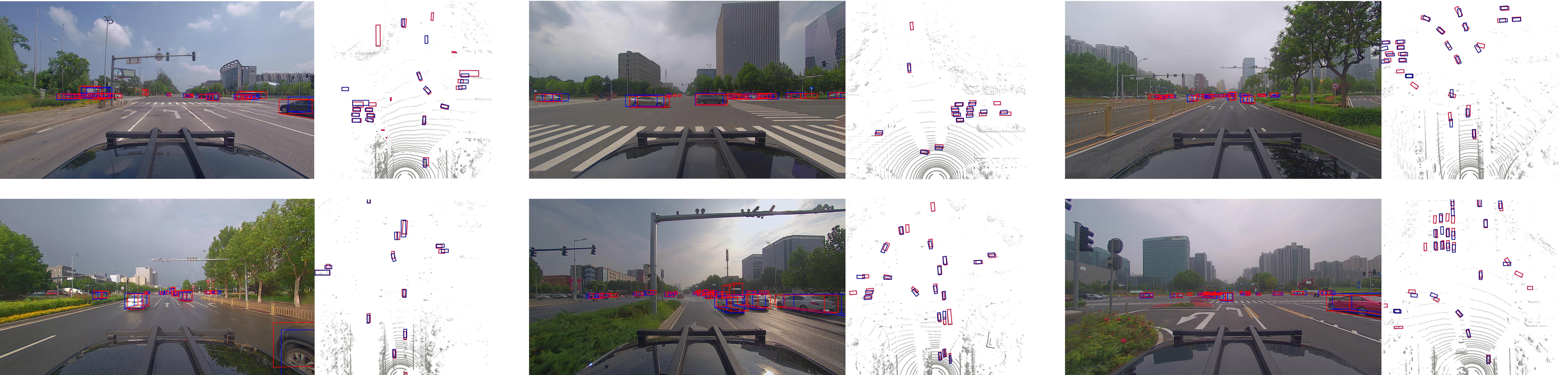}
  \caption{Visualization examples at different scenes. Red: groundtruth. Blue: predictions of QUEST.}
  \label{fig:demo}
\end{figure*}

This section describes experiments on the real-world vehicle-infrastructure dataset. We provide detailed studies and qualitative analysis on effectiveness, flexibility of query transmission, and robustness to packet dropout.

\subsection{Experimental Setting}

\textbf{Datasets.}
We evaluate the proposed QUEST framework on the large-scale real-world cooperative dataset DAIR-V2X-Seq \cite{dair-seq}, which consists of more than 15,000 frames captured from 95 representative scenes. It comprises 7445 image pairs for training and 3316 pairs for validation. We follow the official split scheme and report experimental results on the validation set. The perception range for evaluation is set as $[0, -39, 100, 39]$ following the official setting. The input images are resized to a fixed size of $540 \times 960$.

\begin{figure}[ht]
  \centering
  \includegraphics[scale=0.4]{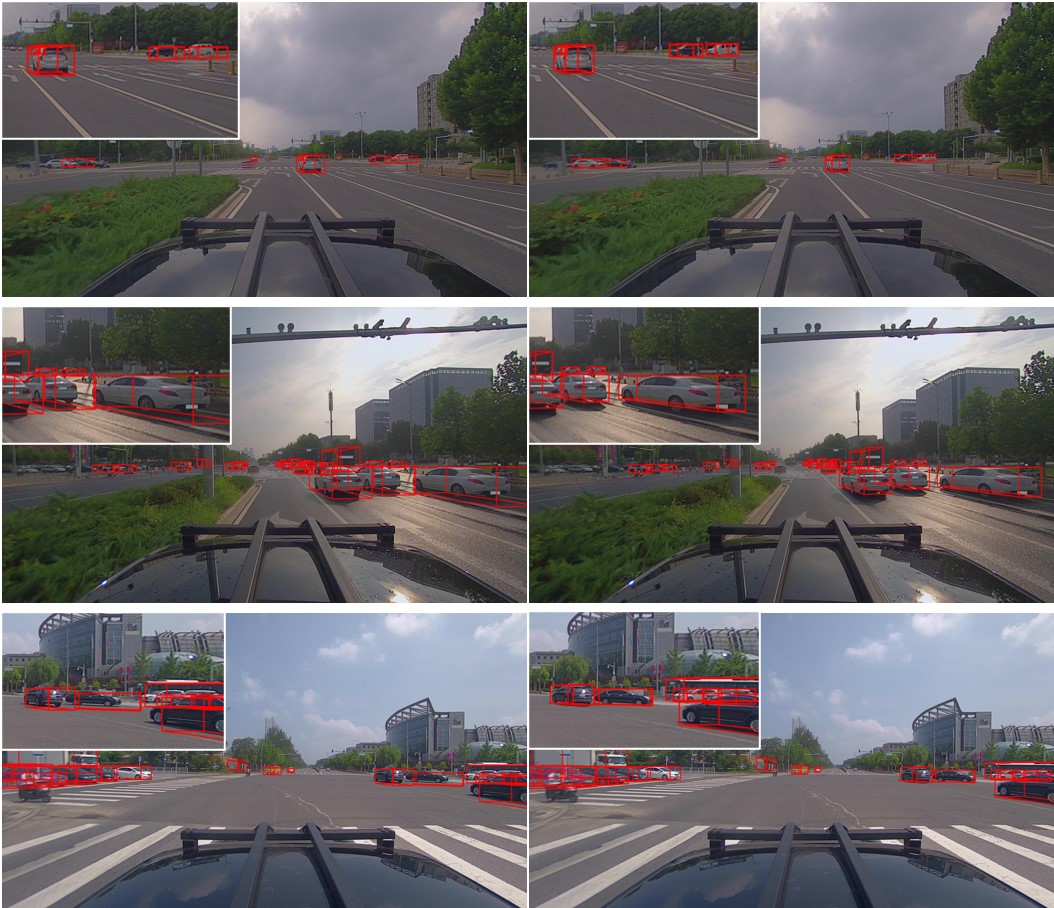}
  \caption{Examples of the generated camera-centric cooperation labels and the corresponding LiDAR-centric labels from \cite{dair-seq}. \textit{Left}: LiDAR-centric labels. \textit{Right}: camera-centric labels. The generated labels will be made publicly available at \href{https://github.com/leofansq/QUEST}{GitHub} upon publication.}
  \label{fig:labels}
\end{figure}

\textbf{Camera-centric cooperation labels.} Since the asynchronous capture frequency between camera and LiDAR, there is always a misalignment between the image and the original cooperation groundtruth (labeled at the LiDAR's timestamp) \cite{yu2022dair}. For the camera-based researches, we generate the cooperation annotations based on the single-side groundtruth labeled at the image-captured timestamps. The generated camera-centric cooperation labels are more accurate, as shown in Figure~\ref{fig:labels}.

\textbf{Implementation Details.}
We employ VoVNetV2 \cite{lee2019energy} as backbone, and the output of the 5th stage is upsampled and fused with that of the 4th stage following PETR \cite{petr}.
AdamW optimizer \cite{loshchilov2017decoupled, adam} is adopted with a weight decay of 0.01. The initial learning rate is set to $2 \times 10^{-4}$ and is scheduled according to cosine annealing \cite{loshchilov2016sgdr}. The model is trained for 100 epochs until convergence. The same as \cite{petr, detr3d, mutr}, the model output at most 300 objects during the inference time. Experiments are implemented in PyTorch on a server with NVIDIA A100.

\subsection{Effectiveness Study}

First of all, we compare our QUEST (two versions) with vehicle-only and result cooperation approaches in Table~\ref{tab:effectiveness}. All reported methods use PETR (adopting VoVNetV2 as backbone) as an individual perception module. The full version QUEST achieves 20.3\% on $AP_{BEV|0.5}$ and 14.1\% on $AP_{3D|0.5}$, which outperforms result cooperation with a large margin, not to mention the vehicle-only approach. Benefiting from the cooperative perception, both distant and occluded objects can be detected, as shown in Figure~\ref{fig:demo}.

Theoretically, there are two ways that query cooperation can boost the perception performance. One is the query enhancement for co-aware objects, the other is the query complementation for unaware objects caused by occlusion or the long-range problem. Therefore, we also report the results of an ablated version (QUEST-f), which only adopts query fusion as cross-agent query interaction, and the query complementation is switched off. 

Noting that QUEST-f performs better than the vehicle-only approach, but is slightly worse than result cooperation. It demonstrates that: (1) If an object can be observed by both vehicle and infrastructure, query fusion can effectively enhance the instance-level feature leveraging the information from another viewpoint; (2) Query complementation is more dominant compared with query fusion, since the unobservable object lies in the blind area of the vehicle can be replenished, which is in line with the motivation of cooperative perception. The instance-level complementation lets result cooperation outperform QUEST-f, but there is a further performance lift when adopting query complementation. Although both of them are at the instance-level, the advantage of query cooperation is more obvious.

\begin{table}[t]
  \small
  \centering
  \caption{Effectiveness study on QUEST framework. QUEST-f is an ablated version that only adopts query fusion (without query complementation) for cross-agent query interaction.}
  \begin{tabular}{c|cc|cc} 
  \hline
    \multirow{2}*{\textbf{Approach}} & \multicolumn{2}{c|}{$AP_{BEV}(\%)$} & \multicolumn{2}{c}{$AP_{3D}(\%)$} \\ 

     & \textbf{$IoU_{0.3}$} & \textbf{$IoU_{0.5}$} & \textbf{$IoU_{0.3}$} & \textbf{$IoU_{0.5}$} \\ \hline

     vehicle-only       & 17.8 & 10.9 & 15.6 & 9.4 \\
     result coop.       & 29.9 & 14.7 & 20.7 & 10.7 \\ \hline
     QUEST-f            & 21.7 & 12.8 & 19.3 & 10.7 \\
     QUEST              & \bf 39.4 & \bf 20.3 & \bf 33.3 & \bf 14.1 \\ \hline
    
  \end{tabular}
  \label{tab:effectiveness}
\end{table}

\subsection{Flexibility of Query Transmission}

Benefiting from the interpretable instance-level cooperation, the cross-agent information transmission is more flexible via query selection. It can be regarded as an instance-level spatial-wise information compression considering wireless bandwidth. QUEST employs confidence-based query selection by filtering the queries under the required score threshold. We report the performance at different thresholds (from 0.1 to 0.8) in Table~\ref{tab:flexiblity}. 

\begin{table}[ht]
  \small
  \centering
  \caption{Performance under different transmission threshold.}
  \begin{tabular}{c|cc|cc|c} 
  \hline
    \multirow{2}*{\textbf{Threshold}} & \multicolumn{2}{c|}{$AP_{BEV}(\%)$} & \multicolumn{2}{c|}{$AP_{3D}(\%)$} & \multirow{2}*{\textbf{Bytes}} \\ 

     & \textbf{$IoU_{0.3}$} & \textbf{$IoU_{0.5}$} & \textbf{$IoU_{0.3}$} & \textbf{$IoU_{0.5}$} & \\ \hline

     0.1 & 40.1 & 20.3 & 33.4 & 14.1 & 74.4K \\
     0.2 & 39.5 & 20.3 & 33.3 & 14.1 & 60.0K \\
     0.3 & 39.4 & 20.3 & 33.3 & 14.1 & 52.2K \\
     0.4 & 39.0 & 20.1 & 33.2 & 14.1 & 43.8K \\
     0.5 & 38.7 & 20.0 & 33.1 & 14.0 & 40.8K \\
     0.6 & 38.3 & 19.7 & 32.5 & 13.8 & 38.2K \\
     0.7 & 37.7 & 19.1 & 32.1 & 13.6 & 35.5K \\
     0.8 & 36.7 & 18.5 & 30.9 & 13.1 & 31.9K \\ \hline
    
  \end{tabular}
  \label{tab:flexiblity}
\end{table}

\begin{figure}[ht]
  \centering
  \includegraphics[scale=0.26]{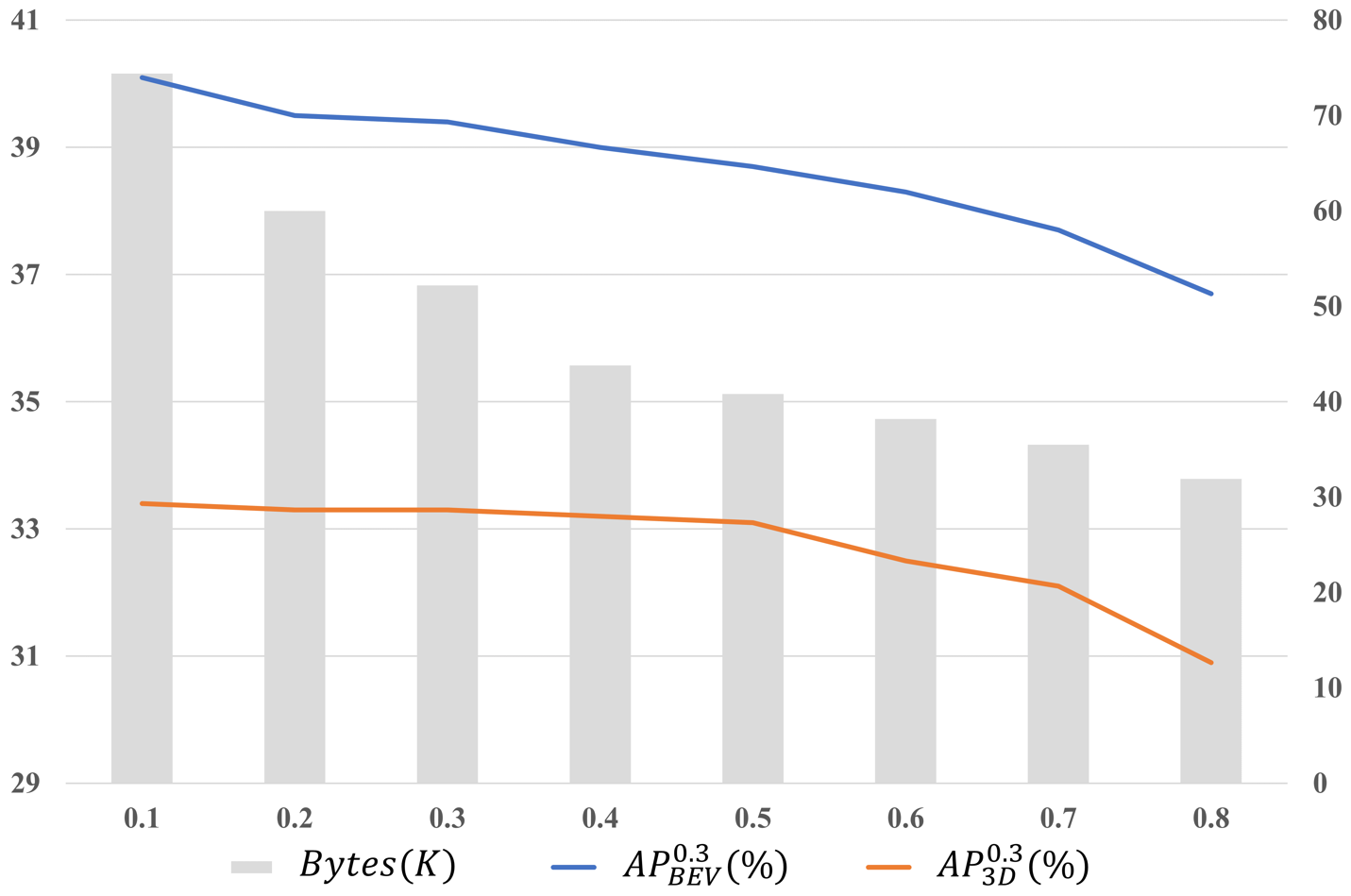}
  \caption{The change of performance and transmission cost under different transmission thresholds.}
  \label{fig:flexible}
\end{figure}

It can be seen that the requirement of transmission bandwidth is significantly reduced as the selection threshold increases (Figure~\ref{fig:flexible}). The transmission Bytes are only half of the full package when we set a higher confidence threshold, such as 0.5. Theoretically, a higher threshold leads to better precision and worse recall. Although both $AP_{BEV}$ and $AP_{3D}$ inevitably decline due to the selection, the descending range is acceptable. 

Compared with region-level spatial-wise compression in the existing feature cooperation approaches, instance-level query selection is more fine-grained and interpretable. The channel-wise query compression can further reduce bandwidth requirements and make it more suitable for practical applications.

\subsection{Robustness to Packet Dropout}

Packet dropout is inevitable for wireless communication, and will severely affect the performance of cooperative perception. 
The scene-level cooperation may degrade into vehicle-side individual perception when the received data/feature is fragmentary due to the packet dropout. Different from that, the minimum transmission unit is reduced to instance level in query cooperation, so the dropout will result in at most partial query loss. 

To simulate the packet dropout, we manually set a dropout ratio of query transmission during evaluation, and the results are reported in Table~\ref{tab:robustness}.

\begin{table}[ht]
  \small
  \centering
  \caption{Performance under different transmission packet dropout ratios. The transmission threshold is set to 0.3.}
  \begin{tabular}{c|cc|cc|c} 
  \hline
    \multirow{2}*{\textbf{Ratio}} & \multicolumn{2}{c|}{$AP_{BEV}(\%)$} & \multicolumn{2}{c|}{$AP_{3D}(\%)$} & \multirow{2}*{\textbf{Bytes}} \\ 

     & \textbf{$IoU_{0.3}$} & \textbf{$IoU_{0.5}$} & \textbf{$IoU_{0.3}$} & \textbf{$IoU_{0.5}$} & \\ \hline
    
     0.0 & 39.4 & 20.3 & 33.3 & 14.1 & 52.2K \\
     0.3 & 33.7 & 17.3 & 28.5 & 12.5 & 36.5K \\
     0.5 & 29.7 & 15.7 & 25.7 & 12.1 & 26.1K \\
     0.7 & 25.9 & 13.7 & 22.2 & 11.6 & 15.6K \\ 
     veh. only & 17.8 & 10.9 & 15.6 & 9.4 & - \\ \hline
    
  \end{tabular}
  \label{tab:robustness}
\end{table}

Although performance decline is avoidless, QUEST can still generate valid predictions when packet dropout occurs. It maintains about $70\%$ performance when the dropout ratio reaches 0.7. The results suggest that QUEST is relatively robust when facing query loss, and show the advantage of query cooperation on robustness to packet dropout.

\section{Discussion on Query Cooperation}

Experimental results of QUEST have reflected the characteristics of query cooperation. In this section, we further discuss the pros and cons of query cooperation paradigm.

\textbf{Possible extensions}. Standing on the midpoint of instance-level result cooperation and scene-level feature cooperation, query cooperation takes both advantages of them, resulting in more possibilities to explore. Since the query stream is instance-level, it is more convenient to introduce temporal information and give the chance to model the individual motion of every single object. Leveraging temporal features, the object detection performance will be further boosted via spatial-temporal cooperation. Similar to single-vehicle scenario, query cooperation paradigm opens the gate to end-to-end (E2E) cooperative tracking via a spatial-temporal query stream. Furthermore, there is a wider ocean to explore, when the query stream goes beyond perception and flows throughout the whole pipeline, including perception, prediction, and planning. E2E cooperative driving can expand the E2E autonomous driving \cite{uniad} to a system-wide improvement for intelligent transportation system.

\textbf{Foreseeable limitation}. Behind all the advances, the limitation is also foreseeable. Since query cooperation is on the basis of the query stream, it naturally requests all agents participating in the symbioses to employ a query-based onboard system. Therefore, the query cooperation adaption for the hybrid intelligent transportation system deserves further exploration. In addition, the query alignment among different transformer-based architectures also needs to be tackled for widespread use. 

\section{Conclusion}

Aiming at interpretable and flexible cooperative perception, we propose the concept of query cooperation in this paper, which enables instance-level feature interaction among agents via the query stream. To specifically describe the query cooperation, a representative cooperative perception framework (QUEST) is proposed. It performs cross-agent query interaction by fusion and complementation, which are designed for co-aware objects and unaware objects respectively. Taking camera-based vehicle-infrastructure cooperative perception as a typical scenario, we generate the camera-centric cooperation labels of DAIR-V2X-Seq and evaluate the proposed framework on it. The experimental results not only demonstrate the effectiveness but also show the advantages of transmission flexibility and robustness to packet dropout. In addition, we discuss the pros and cons of query cooperation paradigm from the possible extensions and foreseeable limitations. 

From our perspective of view, the query cooperation has great potential and deserves further exploration. We hope our work can facilitate the cooperative perception research for practical applications. Planned future efforts will include 1) adaption for other cooperative tasks, e.g., prediction and planning, 2) query alignment across agents and time, and 3) query selection and compression for practical convenience.


\newpage

\bibliographystyle{IEEEtran}
\bibliography{mybib}

\begin{thebibliography}{10}
\providecommand{\url}[1]{#1}
\csname url@rmstyle\endcsname
\providecommand{\newblock}{\relax}
\providecommand{\bibinfo}[2]{#2}
\providecommand\BIBentrySTDinterwordspacing{\spaceskip=0pt\relax}
\providecommand\BIBentryALTinterwordstretchfactor{4}
\providecommand\BIBentryALTinterwordspacing{\spaceskip=\fontdimen2\font plus
\BIBentryALTinterwordstretchfactor\fontdimen3\font minus \fontdimen4\font\relax}
\providecommand\BIBforeignlanguage[2]{{%
\expandafter\ifx\csname l@#1\endcsname\relax
\typeout{** WARNING: IEEEtran.bst: No hyphenation pattern has been}%
\typeout{** loaded for the language `#1'. Using the pattern for}%
\typeout{** the default language instead.}%
\else
\language=\csname l@#1\endcsname
\fi
#2}}

\bibitem{9732063}
A.~Caillot, S.~Ouerghi, P.~Vasseur, R.~Boutteau, and Y.~Dupuis, ``Survey on cooperative perception in an automotive context,'' \emph{IEEE Trans. Intell. Transp. Syst.}, vol.~23, no.~9, pp. 14\,204--14\,223, 2022.

\bibitem{han2023collaborative}
Y.~Han, H.~Zhang, H.~Li, Y.~Jin, C.~Lang, and Y.~Li, ``Collaborative perception in autonomous driving: Methods, datasets and challenges,'' \emph{arXiv preprint arXiv:2301.06262}, 2023.

\bibitem{yu2022dair}
H.~Yu, Y.~Luo, M.~Shu, Y.~Huo, Z.~Yang, Y.~Shi, Z.~Guo, H.~Li, X.~Hu, J.~Yuan, and Z.~Nie, ``{DAIR-V2X}: A large-scale dataset for vehicle-infrastructure cooperative {3D} object detection,'' in \emph{CVPR}, 2022, pp. 21\,361--21\,370.

\bibitem{cbr}
S.~Fan, Z.~Wang, X.~Huo, Y.~Wang, and J.~Liu, ``Calibration-free {BEV} representation for infrastructure perception,'' \emph{arXiv preprint arXiv:2303.03583}, 2023.

\bibitem{bai2022survey}
Z.~Bai, G.~Wu, M.~J. Barth, Y.~Liu, E.~A. Sisbot, K.~Oguchi, and Z.~Huang, ``A survey and framework of cooperative perception: From heterogeneous singleton to hierarchical cooperation,'' \emph{arXiv preprint arXiv:2208.10590}, 2022.

\bibitem{9228884}
E.~Arnold, M.~Dianati, R.~de~Temple, and S.~Fallah, ``Cooperative perception for {3D} object detection in driving scenarios using infrastructure sensors,'' \emph{IEEE Trans. Intell. Transp. Syst.}, vol.~23, no.~3, pp. 1852--1864, 2022.

\bibitem{cooper}
Q.~Chen, S.~Tang, Q.~Yang, and S.~Fu, ``{COOPER}: Cooperative perception for connected autonomous vehicles based on {3D} point clouds,'' in \emph{Proc. Int. Conf. Distrib. Comput. Syst.}, 2019, pp. 514--524.

\bibitem{fcooper}
Q.~Chen, X.~Ma, S.~Tang, J.~Guo, Q.~Yang, and S.~Fu, ``{F-COOPER}: Feature based cooperative perception for autonomous vehicle edge computing system using {3D} point clouds,'' in \emph{Proceedings of the 4th ACM/IEEE Symposium on Edge Computing}, 2019, pp. 88--100.

\bibitem{cui2022coopernaut}
J.~Cui, H.~Qiu, D.~Chen, P.~Stone, and Y.~Zhu, ``{COOPERNAUT}: end-to-end driving with cooperative perception for networked vehicles,'' in \emph{CVPR}, 2022, pp. 17\,252--17\,262.

\bibitem{when2com}
Y.-C. Liu, J.~Tian, N.~Glaser, and Z.~Kira, ``When2com: Multi-agent perception via communication graph grouping,'' in \emph{CVPR}, 2020, pp. 4106--4115.

\bibitem{wang2020v2vnet}
T.-H. Wang, S.~Manivasagam, M.~Liang, B.~Yang, W.~Zeng, and R.~Urtasun, ``{V2VNet}: Vehicle-to-vehicle communication for joint perception and prediction,'' in \emph{ECCV}, 2020, pp. 605--621.

\bibitem{xu2022v2x}
R.~Xu, H.~Xiang, Z.~Tu, X.~Xia, M.-H. Yang, and J.~Ma, ``{V2X-ViT}: Vehicle-to-everything cooperative perception with vision transformer,'' in \emph{ECCV}, 2022, pp. 107--124.

\bibitem{wang2023vimi}
Z.~Wang, S.~Fan, X.~Huo, T.~Xu, Y.~Wang, J.~Liu, Y.~Chen, and Y.-Q. Zhang, ``{VIMI}: Vehicle-infrastructure multi-view intermediate fusion for camera-based {3D} object detection,'' \emph{arXiv preprint arXiv:2303.10975}, 2023.

\bibitem{coca3d}
Y.~Hu, Y.~Lu, R.~Xu, W.~Xie, S.~Chen, and Y.~Wang, ``Collaboration helps camera overtake lidar in {3D} detection,'' in \emph{CVPR}, 2023, pp. 9243--9252.

\bibitem{ffnet}
H.~Yu, Y.~Tang, E.~Xie, J.~Mao, J.~Yuan, P.~Luo, and Z.~Nie, ``Vehicle-infrastructure cooperative 3d object detection via feature flow prediction,'' \emph{arXiv preprint arXiv:2303.10552}, 2023.

\bibitem{where2comm}
Y.~Hu, S.~Fang, Z.~Lei, Y.~Zhong, and S.~Chen, ``Where2comm: Communication-efficient collaborative perception via spatial confidence maps,'' \emph{arXiv preprint arXiv:2209.12836}, 2022.

\bibitem{detr}
N.~Carion, F.~Massa, G.~Synnaeve, N.~Usunier, A.~Kirillov, and S.~Zagoruyko, ``End-to-end object detection with transformers,'' in \emph{ECCV}, 2020, pp. 213--229.

\bibitem{detr3d}
Y.~Wang, V.~C. Guizilini, T.~Zhang, Y.~Wang, H.~Zhao, and J.~Solomon, ``{DETR3D}: {3D} object detection from multi-view images via {3D-to-2D} queries,'' in \emph{CoRL}, 2022, pp. 180--191.

\bibitem{motr}
F.~Zeng, B.~Dong, Y.~Zhang, T.~Wang, X.~Zhang, and Y.~Wei, ``{MOTR}: End-to-end multiple-object tracking with transformer,'' in \emph{ECCV}, 2022, pp. 659--675.

\bibitem{petr}
Y.~Liu, T.~Wang, X.~Zhang, and J.~Sun, ``{PETR}: Position embedding transformation for multi-view {3D} object detection,'' in \emph{ECCV}, 2022, pp. 531--548.

\bibitem{bevformer}
Z.~Li, W.~Wang, H.~Li, E.~Xie, C.~Sima, T.~Lu, Y.~Qiao, and J.~Dai, ``{BEVFormer}: Learning bird’s-eye-view representation from multi-camera images via spatiotemporal transformers,'' in \emph{ECCV}, 2022, pp. 1--18.

\bibitem{bevformerv2}
C.~Yang, Y.~Chen, H.~Tian, C.~Tao, X.~Zhu, Z.~Zhang, G.~Huang, H.~Li, Y.~Qiao, L.~Lu, J.~Zhou, and J.~Dai, ``{BEVFormer} v2: Adapting modern image backbones to bird's-eye-view recognition via perspective supervision,'' in \emph{CVPR}, 2023, pp. 17\,830--17\,839.

\bibitem{trackformer}
T.~Meinhardt, A.~Kirillov, L.~Leal-Taixe, and C.~Feichtenhofer, ``Trackformer: Multi-object tracking with transformers,'' in \emph{CVPR}, 2022, pp. 8844--8854.

\bibitem{mutr}
T.~Zhang, X.~Chen, Y.~Wang, Y.~Wang, and H.~Zhao, ``{MUTR3D}: A multi-camera tracking framework via {3D-to2D} queries,'' in \emph{CVPRW}, 2022, pp. 4537--4546.

\bibitem{pftr}
Z.~Pang, J.~Li, P.~Tokmakov, D.~Chen, S.~Zagoruyko, and Y.-X. Wang, ``Standing between past and future: Spatio-temporal modeling for multi-camera {3D} multi-object tracking,'' in \emph{CVPR}, 2023, pp. 17\,928--17\,938.

\bibitem{dair-seq}
H.~Yu, W.~Yang, H.~Ruan, Z.~Yang, Y.~Tang, X.~Gao, X.~Hao, Y.~Shi, Y.~Pan, N.~Sun, J.~Song, J.~Yuan, P.~Luo, and Z.~Nie, ``{V2X-Seq}: A large-scale sequential dataset for vehicle-infrastructure cooperative perception and forecasting,'' in \emph{CVPR}, 2023, pp. 5486--5495.

\bibitem{lee2019energy}
Y.~Lee, J.-W. Hwang, S.~Lee, Y.~Bae, and J.~Park, ``An energy and gpu-computation efficient backbone network for real-time object detection,'' in \emph{CVPRW}, 2019, pp. 0--0.

\bibitem{loshchilov2017decoupled}
I.~Loshchilov and F.~Hutter, ``Decoupled weight decay regularization,'' \emph{arXiv preprint arXiv:1711.05101}, 2017.

\bibitem{adam}
D.~P. Kingma and J.~Ba, ``Adam: A method for stochastic optimization,'' \emph{arXiv preprint arXiv:1412.6980}, 2014.

\bibitem{loshchilov2016sgdr}
I.~Loshchilov and F.~Hutter, ``{SGDR}: Stochastic gradient descent with warm restarts,'' \emph{arXiv preprint arXiv:1608.03983}, 2016.

\bibitem{uniad}
Y.~Hu, J.~Yang, L.~Chen, K.~Li, C.~Sima, X.~Zhu, S.~Chai, S.~Du, T.~Lin, W.~Wang, L.~Lu, X.~Jia, Q.~Liu, J.~Dai, Y.~Qiao, and H.~Li, ``Planning-oriented autonomous driving,'' in \emph{CVPR}, 2023, pp. 17\,853--17\,862.

\end{thebibliography}

\end{document}